\documentclass[10pt,twocolumn,letterpaper]{article}
\usepackage[numbers, sort, compress]{natbib} 

\usepackage{iccv}
\usepackage{times}
\usepackage{epsfig}
\usepackage{graphicx,graphics}
\usepackage{amsmath}
\usepackage{amssymb}
\usepackage{booktabs}

\usepackage{multirow}
\usepackage{multicol}
\usepackage[american]{babel}
\usepackage{microtype}
\usepackage{booktabs,bigstrut}
\usepackage{enumitem}
\usepackage{caption}
\usepackage{algorithm}
\usepackage{algorithmic}
\usepackage{framed}
\usepackage{amsthm}
\usepackage{color}
\usepackage{graphicx}
\usepackage{arydshln}

\usepackage{subcaption}

\graphicspath{{Figure/}}

\usepackage{color, xcolor, colortbl}
\definecolor{citecolor}{HTML}{229954}
\usepackage[pagebackref=true,breaklinks=true,colorlinks,citecolor=citecolor,bookmarks=false]{hyperref}

\newcommand{\red}[1]{\textcolor{red}{#1}}

\newcommand{\bfsection}[1]{\vspace*{0.1cm}\noindent\textbf{#1.}}

\usepackage[capitalize]{cleveref}
\crefname{section}{Sec.}{Secs.}
\Crefname{section}{Section}{Sections}
\Crefname{table}{Table}{Tables}
\crefname{table}{Tab.}{Tabs.}

\theoremstyle{definition}
\newtheorem{definition}{Definition}

\makeatletter
\renewcommand{\maketag@@@}[1]{\hbox{\m@th\normalsize\normalfont#1}}%
\makeatother

\iccvfinalcopy 

\def\iccvPaperID{3569} 

\ificcvfinal\fi

\begin{document}

    \title{Spherical Space Feature Decomposition for Guided Depth Map Super-Resolution}
    \author{Zixiang Zhao$^{1,2}$\quad
            Jiangshe Zhang$^{1}$\thanks{Corresponding author.}\quad
            Xiang Gu$^{1}$\quad
            Chengli Tan$^{1}$\\
            Shuang Xu$^{3}$\quad
            Yulun Zhang$^{2}$\quad
            Radu Timofte$^{2,4}$\quad
            Luc Van Gool$^{2}$\\[2mm]
            $^{1}$Xi’an Jiaotong University \quad
            $^{2}$Computer Vision Lab, ETH Z\"urich\\
            $^{3}$Northwestern Polytechnical University\quad
            $^{4}$University of W\"urzburg\\
            {\tt\small zixiangzhao@stu.xjtu.edu.cn, jszhang@mail.xjtu.edu.cn}
    }

    \maketitle
    \ificcvfinal\thispagestyle{empty}\fi

    \begin{abstract}
        Guided depth map super-resolution (GDSR), as a hot topic in multi-modal image processing, aims to upsample low-resolution (LR) depth maps with additional information involved in high-resolution (HR) RGB images from the same scene. The critical step of this task is to effectively extract domain-shared and domain-private RGB/depth features. In addition, three detailed issues, namely blurry edges, noisy surfaces, and over-transferred RGB texture, need to be addressed. In this paper, we propose the Spherical Space feature Decomposition Network (SSDNet) to solve the above issues. To better model cross-modality features, Restormer block-based RGB/depth encoders are employed for extracting local-global features. Then, the extracted features are mapped to the spherical space to complete the separation of private features and the alignment of shared features. Shared features of RGB are fused with the depth features to complete the GDSR task. Subsequently, a spherical contrast refinement (SCR) module is proposed to further address the detail issues. Patches that are classified according to imperfect categories are input into the SCR module, where the patch features are pulled closer to the ground truth and pushed away from the corresponding imperfect samples in the spherical feature space via contrastive learning. Extensive experiments demonstrate that our method can achieve state-of-the-art results on four test datasets, as well as successfully generalize to real-world scenes. The code is available at \url{https://github.com/Zhaozixiang1228/GDSR-SSDNet}.
    \end{abstract}
    \newcommand{\parameterimagewidth}{0.49}
    \begin{figure}[t]
        \centering
        \begin{subfigure}{\parameterimagewidth\linewidth}
            \centering
            \includegraphics[width=\linewidth]{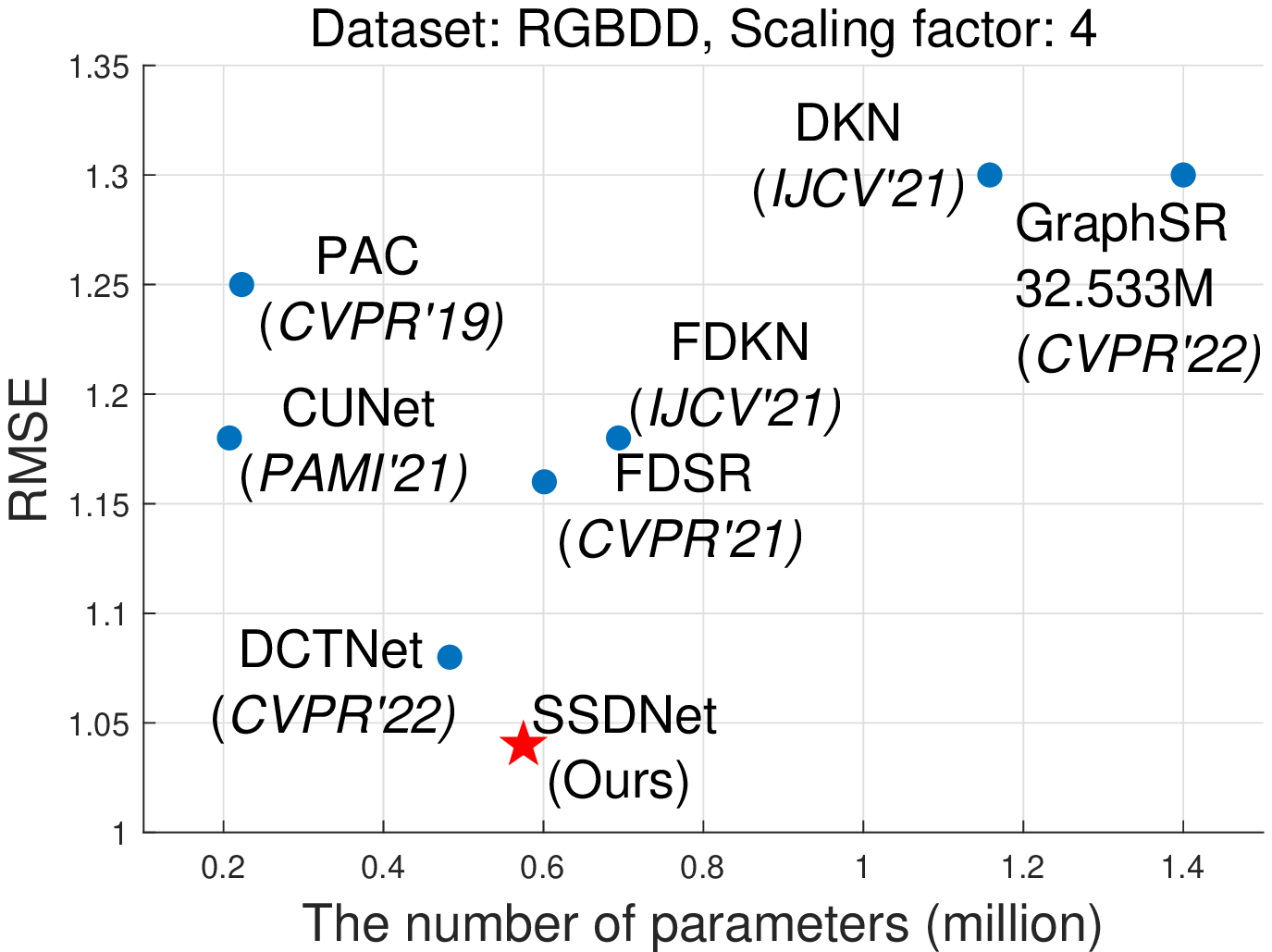}
        \end{subfigure}
        \begin{subfigure}{\parameterimagewidth\linewidth}
            \centering
            \includegraphics[width=\linewidth]{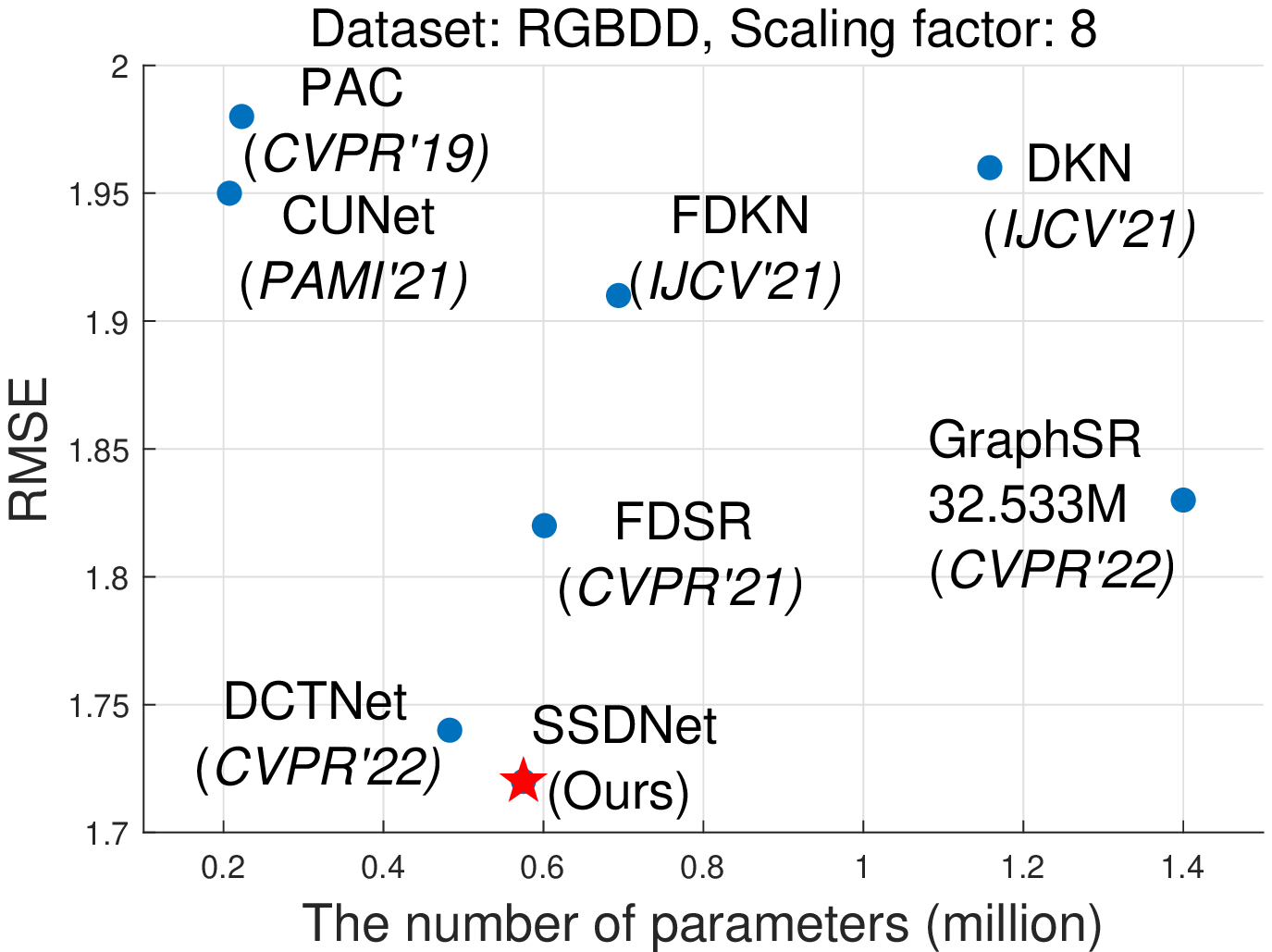}
        \end{subfigure}
        \begin{subfigure}{\parameterimagewidth\linewidth}
            \centering
            \includegraphics[width=\linewidth]{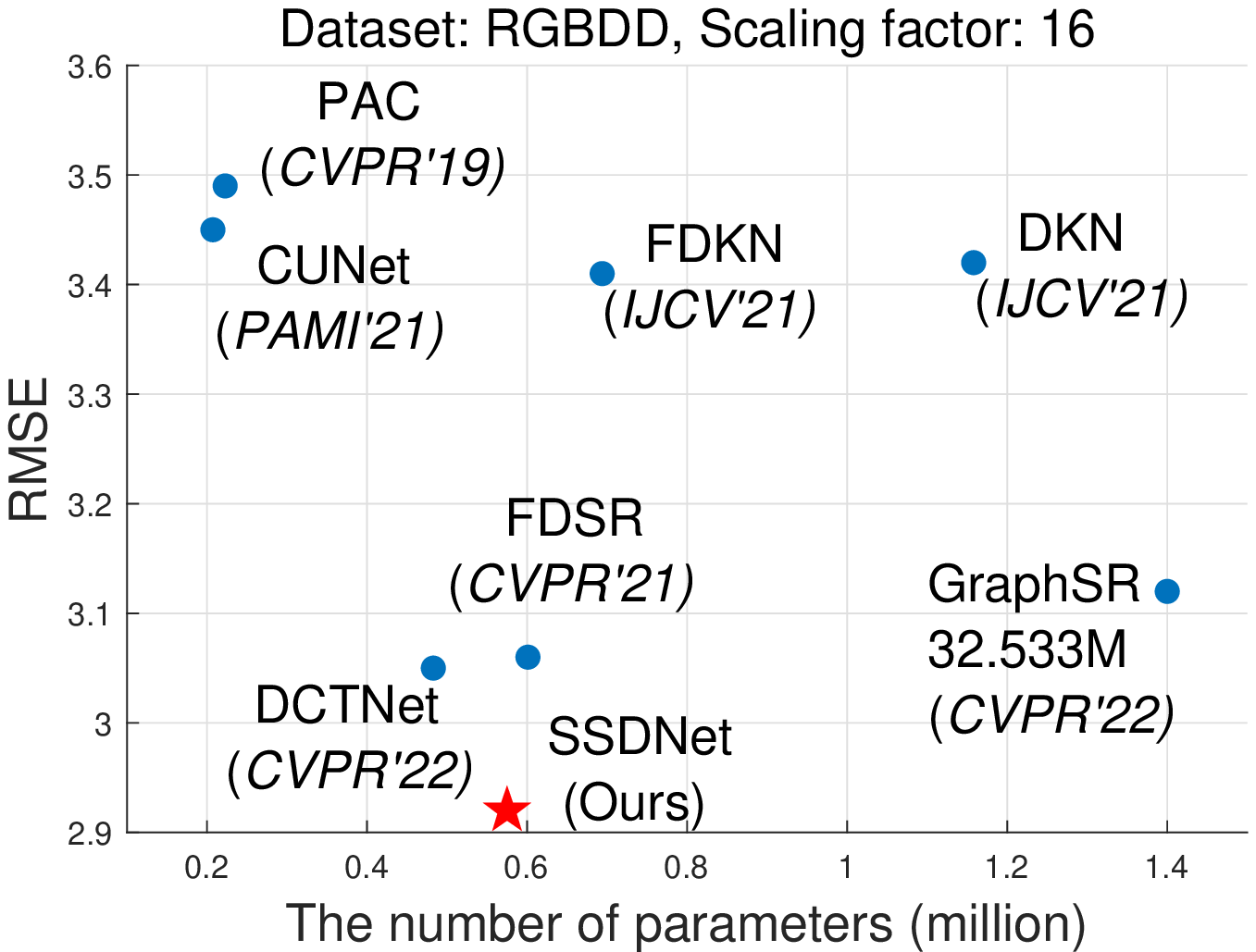}
        \end{subfigure}
        \begin{subfigure}{\parameterimagewidth\linewidth}
            \centering
            \includegraphics[width=\linewidth]{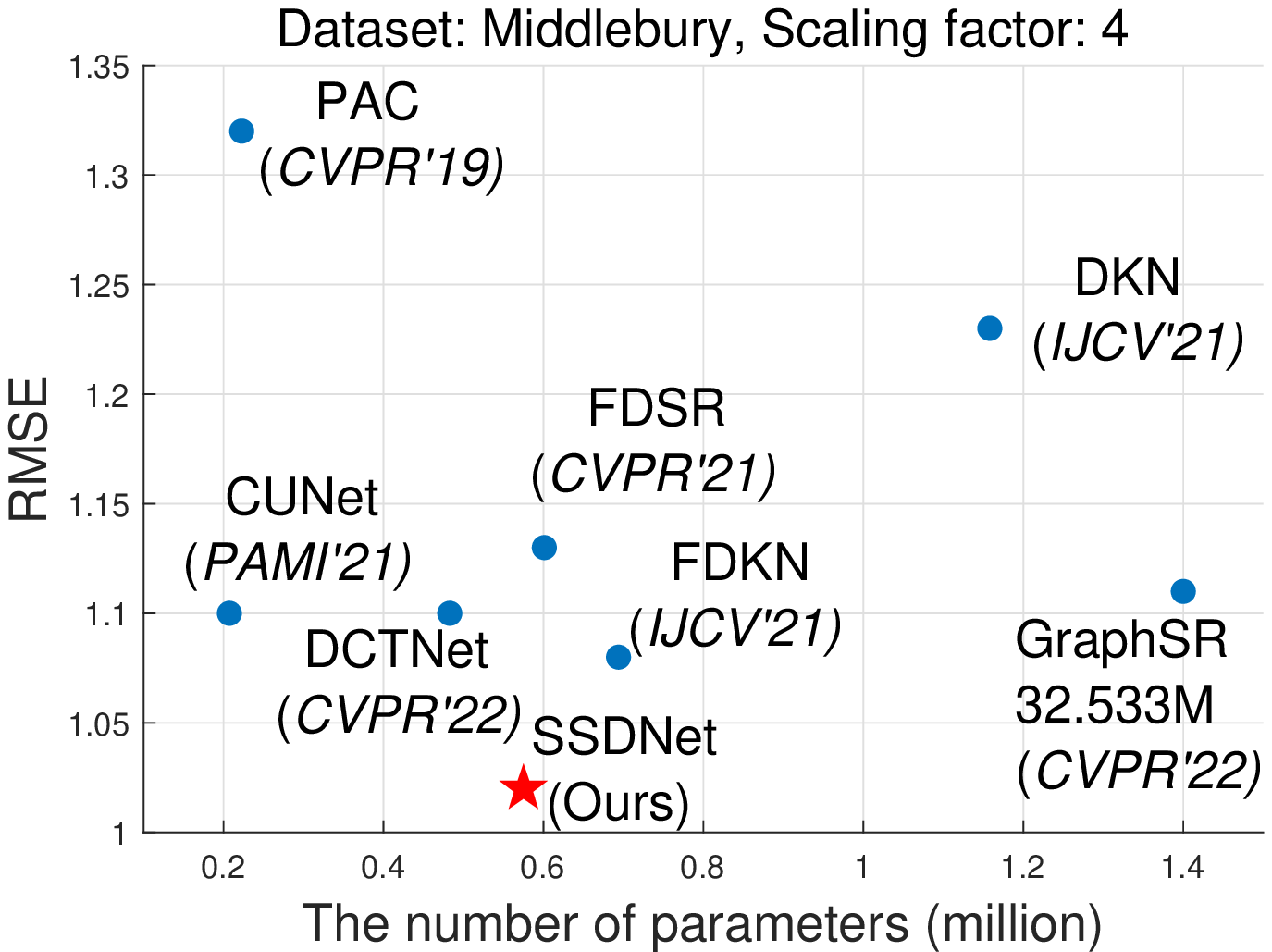}
        \end{subfigure}
        \vspace{-0.1em}
        \caption{Our SSDNet achieves outstanding performance on the RGBDD dataset for $\times$4, $\times$8, and $\times$16 and Middlebury for $\times$4 while being computationally efficient.}
        \label{fig:introduction}
    \end{figure}
    \section{Introduction}\label{sec:1}
    Depth maps, as images that measure the distance of scene points from the sensor, are widely used in autonomous driving~\cite{DBLP:conf/eccv/LiaoLZZLY20,DBLP:conf/cvpr/PengPLS20,zhao2023equivariant,DBLP:journals/corr/abs-2211-14461,DBLP:journals/corr/abs-2303-06840}, pose estimation~\cite{DBLP:conf/cvpr/YuanGSMCLMKHGYC18,DBLP:conf/iccv/XiongZ0CYZY19}, virtual reality~\cite{DBLP:journals/tip/LiuZCJZG19a,qinbipointnet,yao2023towards,yao2023bidirectional}, and scene understanding~\cite{DBLP:journals/tog/IzadiF14,DBLP:conf/eccv/GuptaGAM14,wang2021universal,qin2020forward}.
    However, the depth maps produced by current consumer-level depth sensors, \eg, Time-of-Flight~(ToF) and Kinect cameras, often have the disadvantages of low resolution and noise. These disadvantages are insufficient to meet the requirements of advanced computing vision tasks~\cite{zhong2023guided,qin2020binary,xu2023zero}.

    To obtain high-resolution depth maps, we naturally hope to accomplish depth map super-resolution (SR) by transferring mature SR models in the RGB image domain~\cite{DBLP:conf/eccv/DongLHT14,lim2017enhanced,DBLP:conf/cvpr/ZhangGT20} to the depth SR task.
    However, a potential risk is that RGB SR models tend to focus on reconstructing high-frequency image information, such as details and texture.
    Conversely, for depth images, the objects' depth information is often textureless and piecewise, and more sensitive to unclear edges and noise.
    Therefore, it is unreasonable to directly apply RGB SR models to the depth SR task.
    On the other hand, while acquiring the depth map, it is relatively easy to obtain HR and noise-free RGB images in the same scene. Furthermore, there are statistical co-occurrences between the edges in RGB images and the discontinuities in depth maps~\cite{DBLP:conf/eccv/RieglerRB16}.
    Therefore, we hope to use the HR RGB images to provide edges and contour information missing in the depth map, and to fuse the multi-modal information to accomplish the LR depth image upsampling.

    In the era of deep learning~(DL), numerous methods have been utilized to learn the mapping between LR $\to$ HR depth maps. These methods succeeded to some extent in modeling the cross-modality features and reconstructing the contour and edge information.
    However, DL models which rely on natural priors will limit the model's flexibility. Models learning the LR $\to$ HR mapping via data-driven methods are difficult to interpret due to the unclear working mechanism~\cite{DBLP:conf/cvpr/ZhaoZXLP22}. Thus, effectively extracting and distinguishing domain-shared and domain-private RGB/depth features is still a challenge. While microscopically, the obtained depth images are often plagued by three detail issues: blurry edges, noisy surfaces, and over-transferred RGB texture, all of which affect the display quality of depth maps.

    In response to addressing the above challenges, we expect to limit the solution space by constraining the extracted features and further improve the modeling of the dependencies between different modalities.
    Based on our observation, RGB images and depth maps contain shared features, such as depth map discontinuities and edge features of RGB objects, which can be aligned in the feature space. In addition, unique private features, such as the distance information of the depth map and the texture of RGB object surface, should be separated conversely.
    Thus, in the feature space, domain-shared and domain-private features are expected to separate and align  respectively, while the distance between HR features and imperfect features with the above-mentioned detail issues should also be pushed away.

    The above goal can be divided into two steps. First, extracting features through an effective encoder, and then selecting an appropriate distance measure to complete the alignment and separation of features.
    Considering that CNN-based architectures limit feature extraction capabilities due to content-independent convolution kernels and the lack of global information, we employ the Restormer blocks~\cite{DBLP:conf/cvpr/ZamirA0HK022}, which has been proved to effectively extract features in the low-level vision domain, as the basic unit of our encoder.
    For the choice of distance measure, we first think of the Euclidean distance, such as $\ell_1$ or $\ell_2$ distance. However, cross-modality features often contain different scales and orders of magnitude, and the Euclidean distance is easily affected by the scale, which makes it difficult to achieve our goal.
    Recently, with the development of spherical DL models~\cite{gu2020spherical,9733209,castillo2012geometric,2018spherical}, the spherical feature learning is well known by virtue of its advantages over the Euclidean feature learning in many applications, \eg, domain adaptation~\cite{gu2020spherical}. Due to the distances between spherical features being regularized, the alignment and separation of features can be done more easily without losing the model's representational capacity.

    According to the above analysis, we propose our \textit{Spherical Space feature Decomposition Network} (SSDNet). The specific workflow is displayed in \cref{fig:Workflow}. Our contribution can be summarized in three-fold:
    \begin{itemize}[itemsep=0cm,topsep=0cm,parsep=0pt]
        \item We propose a spherical space feature decomposition framework to model the cross-modality features. The features extracted by the Restormer block are mapped to the spherical space for separation and alignment. This is the first time that the Transformer and spherical-space distance measure are applied to the GDSR task.
        \item The spherical contrast refinement module, cooperating with the imperfect patch classification and the corresponding contrastive learning branch, is proposed to address the possible detail issues in depth maps. This is also the first time that the contrastive learning technique has been used for the GDSR task.
        \item Experiments on four GDSR benchmarks and a real scene dataset demonstrate that our method can generate satisfactory HR depth maps in different scenarios and exhibits good generalization ability.
    \end{itemize}

    \section{Related Work}\label{sec:2}
    In this section, we briefly introduce the GDSR task, and illustrate the Vision Transformer, spherical space DL and contrastive learning techniques utilized in SSDNet.
    \subsection{GDSR methods}
    Image super-resolution is a popular image processing and computer vision task with too many sub-categories, so we only discuss the GDSR task here.
    Conventional GDSR methods can be divided into local-~\cite{DBLP:conf/cvpr/0001TT13,DBLP:journals/tip/MinLD12,DBLP:conf/cvpr/LuSMLD12,DBLP:conf/cvpr/LuF15}, global-~\cite{DBLP:conf/nips/DiebelT05,DBLP:conf/iccv/ParkKTBK11,DBLP:journals/tip/YangYLHW14,DBLP:conf/iccv/FerstlRRRB13,DBLP:conf/eccv/LiMDL16} and learning-based~\cite{DBLP:journals/tmm/XieFYS15,DBLP:conf/cvpr/GuZGCCZ17,gu2019learned,DBLP:journals/tip/XieFS16,DBLP:conf/cvpr/LutioBDRWS22} methods, \etc. They extract cross-modality information by manually designed filters, optimizing equations, or sparse dictionary learning.
    With the rapid development of DL, GDSR is further promoted by the CNN-based methods~\cite{DBLP:journals/tip/YeSWYXLL20,DBLP:journals/tip/WenSLLF19,DBLP:conf/cvpr/SongDZLLLY20,DBLP:conf/cvpr/SunYLLW021,DBLP:conf/mm/TangCSHZZK21,DBLP:conf/mm/TangCZ21}. Generally, DL-based methods can be categorized in three groups, \ie, auto-encoder (AE)~\cite{DBLP:journals/pami/LiHAY19,DBLP:conf/eccv/HuiLT16,DBLP:journals/tip/GuoLGCFH19,DBLP:journals/tip/ZhongLJZCJ22,DBLP:conf/cvpr/ZhaoZXLP22,DBLP:journals/ijcv/YangCZT22}, learnable filter~\cite{DBLP:conf/cvpr/Wu0ZH18,DBLP:journals/ijcv/KimPH21,DBLP:journals/pami/DongPRLTY22}, algorithm unfolding~\cite{DBLP:journals/tip/DengD20,DBLP:conf/bmvc/RieglerFRB16,DBLP:journals/pami/0002D21,DBLP:journals/ijcv/ZhouYPRXC23} and unsupervised~\cite{DBLP:conf/iccv/LutioDWS19,Dong2022MMSR} methods.
    AE-based methods learn cross-modality features via shared or private encoders and reconstruct HR depth maps with decoders.
    In the learnable filter group, the convolution kernels used to extract information are set to be context-dependent and spatially-variant, which improves the flexibility of the model.
    The algorithmic unfolding group establishes interpretable GDSR models by building a bridge between traditional optimization functions and deep learning. Unsupervised methods change the training paradigm that requires paired LR/HR images, thus solving the difficulty of obtaining paired training data.

    \subsection{Spherical deep learning}
    The works~\cite{castillo2012geometric,2018spherical,cohen2017convolutional} introduce the geometrical neural network or the spherical convolution neural networks for analyzing the spherical signals that are rotation invariant. Gu \etal~\cite{9733209} propose the spherical multi-layer perceptron for aligning feature distributions of different domains in spherical feature space for domain adaptation. Our idea of spherical space feature decomposition is inspired by Gu \etal~\cite{9733209} that shows the advantages of spherical feature learning over the Euclidean feature learning for domain adaptation in the ``high-level'' image classification task. Different from~\cite{9733209}, we tackle the ``low-level'' GDSR task by aligning/separating the shared/private features of paired RGB image and depth map, rather than aligning the feature distributions as in~\cite{9733209}. We further propose the spherical contrast refinement for tackling the issues of blurry edges, noisy surfaces, and over-transferred RGB texture for GDSR.
    \subsection{Transformer in vision}
    Transformer, proposed by Vaswani et al. \cite{DBLP:conf/nips/VaswaniSPUJGKP17}, has now become a popular technology in computer vision. It has achieved dominance in  classification~\cite{DBLP:conf/iclr/DosovitskiyB0WZ21,DBLP:conf/icml/TouvronCDMSJ21,DBLP:conf/iccv/LiuL00W0LG21}, object detection \cite{DBLP:conf/eccv/CarionMSUKZ20,DBLP:conf/iclr/ZhuSLLWD21}, segmentation \cite{DBLP:conf/iccv/WangX0FSLL0021,DBLP:conf/cvpr/ZhengLZZLWFFXT021}, \etc.
    Simultaneously, it has also made progress in low-level vision~\cite{DBLP:journals/corr/abs-2106-06847,DBLP:journals/corr/abs-2203-14186,DBLP:journals/corr/abs-2106-03106,DBLP:journals/corr/abs-2112-10175}.
    Chen \etal~\cite{DBLP:conf/cvpr/Chen000DLMX0021} proposed IPT based on the standard Transformer and multi-task learning. Liang \etal~\cite{DBLP:conf/iccvw/LiangCSZGT21} proposed SwinIR whose core module inherits from Swin Transformer~\cite{DBLP:conf/iccv/LiuL00W0LG21}. Recently, Restormer~\cite{DBLP:conf/cvpr/ZamirA0HK022} improves the transformer block by the gated-Dconv network and multi-Dconv head transposed attention.
    By applying self-attention in the feature dimension, the ability of local-global representation learning is maintained while being more friendly to high-resolution input images.
    \subsection{Contrastive Learning}
    Contrastive learning, which aims to address the scarcity of source and target paired data and prioritize self-supervised representation learning~\cite{DBLP:conf/cvpr/He0WXG20,DBLP:conf/icml/Henaff20}.
    The primary objective is to bring samples close to their positive counterparts and push them away from negative ones within high-dimensional manifolds, and it seeks to increase the distance between samples belonging to different classes.
    Recently, contrastive learning has gained significant attention in computer vision, particularly in high-level tasks such as object detection~\cite{DBLP:conf/iccv/XieDWZXSL021}, video segmentation~\cite{DBLP:conf/iccv/CaronTMJMBJ21}, and object tracking~\cite{DBLP:conf/cvpr/YuLH22}. It has also found applications in low-level tasks such as super-resolution~\cite{DBLP:journals/corr/abs-2107-00708}, image translation~\cite{DBLP:conf/eccv/ParkEZZ20}, and image dehazing~\cite{DBLP:conf/cvpr/WuQLZQZ0M21}.
    \subsection{Comparison with existing approaches}
    The methods most related to our approach are feature-decomposition DL methods~\cite{DBLP:journals/tip/DengD20,DBLP:conf/cvpr/ZhaoZXLP22}, which propose the idea that cross-modality features contain the common and private information.
    In contrast, our method, for the first time, exploits the Transformer model's powerful ability to model globally dependent features. Meanwhile, the distance measure in spherical space is proved to decompose the cross-modality features more effectively than Euclidean distances.
    Additionally, instead of only focusing on optimizing the $\ell_2$ loss between the reconstructed image and ground truth, we give classification-based fine-tuning for image patches under different imperfect conditions, thus further improving the performance of GDSR.

    \begin{figure*}[t]
        \centering
        \includegraphics[width=\linewidth]{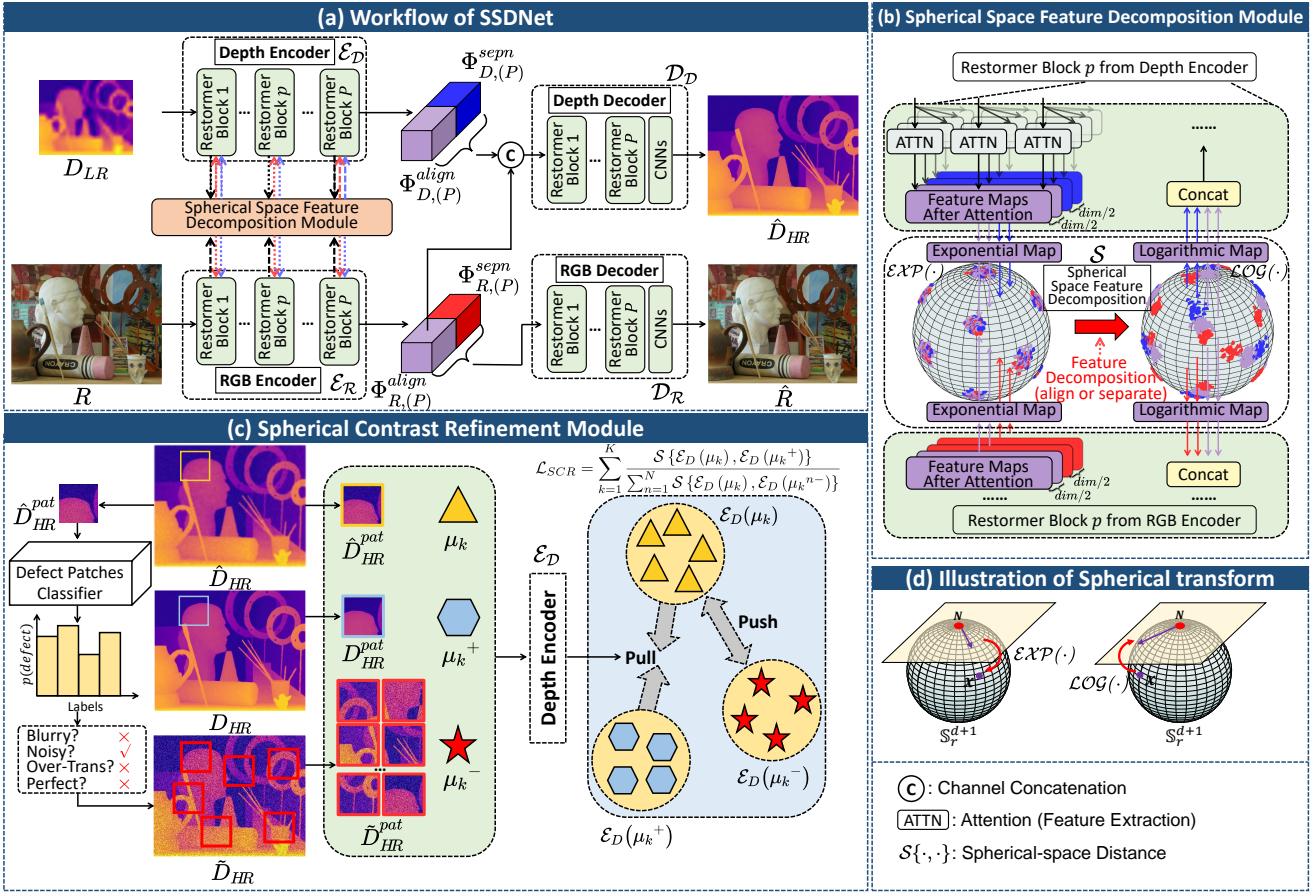}
        \caption{\textbf{(a)} The Architecture of our SSDNet method. First, the LR depth map is input into $\mathcal{E_D}$, and the RGB image is input into $\mathcal{E_R}$. The intermediate feature maps are projected onto the spherical space to complete the feature decomposition, \ie, alignment and separation. The shared \& private features of the depth map, and the shared features of the RGB image are input into the $\mathcal{D_D}$ to reconstruct HR depth map. The RGB features are fed into $\mathcal{D_R}$ to reconstruct RGB image.
        \textbf{(b)} Illustration of spherical space feature decomposition.
        \textbf{(c)} The module for spherical contrast refinement. \textbf{(d)} The mapping between {Euclidean Space} and  {Spherical Space}.}
        \label{fig:Workflow}
    \end{figure*}

    \section{Method}\label{sec:3}
    In this section, we first present the specific architecture of our SSDNet. Then, comprehensive descriptions are provided for each module. Finally, the loss function and training details will also be addressed.

    \subsection{Overview}
    We can define the input LR depth map, RGB image and the HR depth map as $D_{LR}\!\in\!\mathbb{R}^{h\times w}$, $D_{HR}\!\in\!\mathbb{R}^{H\times W}$ and $R\!\in\!\mathbb{R}^{H\times W\times 3}$, where $\{H,W\}$ and $\{h,w\}$ are the height and width of input RGB and depth images, respectively.
    For our SSDNet network, it consists of five modules, namely, the \textit{Encoders} for RGB and Depth images, the \textit{Decoders} for RGB and Depth images, and the \textit{Spherical Contrast Refinement}~(SCR) module, which is denoted as $\mathcal{E_R}(\cdot)$, $\mathcal{E_D}(\cdot)$, $\mathcal{D_R}(\cdot)$, $\mathcal{D_D}(\cdot)$, and $\mathcal{S}(\cdot)$, respectively.

    In general, as shown in \cref{fig:Workflow}, SSDNet utilizes $\mathcal{E_R}$ and $\mathcal{E_D}$ to extract features, which are then projected onto the spherical space for feature separation and alignment. Then, the depth features and shared RGB features are input into $\mathcal{D_R}$ to obtain reconstructed depth maps, and RGB features are fed into $\mathcal{E_R}$ to get reconstructed RGB images. Finally, the reconstructed depth maps are subsequently provided to the SCR module to refine details and output the final $D_{HR}$.

    The basic unit we employed for feature extraction and image restoration is the Restormer block~\cite{DBLP:conf/cvpr/ZamirA0HK022}. The reason we opted for the Restormer block in $\mathcal{E_R}$, $\mathcal{E_D}$, $\mathcal{D_R}$ and $\mathcal{D_D}$ is because it allows for the extraction of global features from high-resolution input images by utilizing self-attention across feature dimensions~\cite{DBLP:conf/cvpr/ZamirA0HK022}. This approach enables the extraction of cross-modality shallow features without significantly increasing computational requirements. For details on the Restormer block architecture, please refer to the supplementary material or the original paper~\cite{DBLP:conf/cvpr/ZamirA0HK022}.

    \subsubsection{Spherical space transform}
    We first define the mapping based on Riemannian geometry between \textit{Euclidean Space} feature maps and \textit{Spherical Space} feature maps, \ie, $\mathcal{LOG}(\cdot)$ and $\mathcal{EXP}(\cdot)$, which are employed in the feature decomposition. A schematic diagram for the mappings is shown in \cref{fig:Workflow}\red{d}. The distance measure on the spherical space is also defined.
    \begin{definition}[Spherical exponential mapping]Given a vector $v$ in $d$-dimensional Euclidean Space, we define the ($d+1$)-dimensional vector $\bar{v}$ by $\bar{v}=(v,r)$ where $r$ is a hyper-parameter named radius. Then the spherical exponential transform \cite{DBLP:journals/pami/WilsonHPD14} $\exp_N:T_N\mathbb{S}^{d+1}_r\rightarrow \mathbb{S}^{d+1}_r$ is defined as
        \begin{equation}
            \exp_N(v) = N\cos{\theta} + \bar{v}\frac{\sin{\theta}}{\theta},
        \end{equation}
        where $\mathbb{S}^{d+1}_r = \{x\in\mathbb{R}^{d+1}:\Vert x\Vert = r\} \subset \mathbb{R}^{d+1}$ is the $d$-dimensional spherical space, $N=(0,\cdots,0,r)\in\mathbb{S}^{d+1}_r$ is the north pole, $\theta=\frac{\Vert \bar{v}\Vert}{r}$, and $T_N\mathbb{S}^{d+1}_r = \{(v,r):v\in\mathbb{R}^d\}$ is the tangent space of  $\mathbb{S}^{d+1}_r$.
        Further, the spherical exponential mapping $\mathcal{EXP}:\mathbb{R}^{h'\times w'\times d}\rightarrow\mathbb{R}^{h'\times w'\times (d+1)}$ for an Euclidean feature map $\Phi$ is defined as
        \begin{equation}
            \mathcal{EXP}(\Phi)[i,j,:] = \exp_N(\Phi[i,j,:])
        \end{equation}
        where $\Phi[i,j,:]$ is the feature in location $(i,j)$.
    \end{definition}
    \begin{definition}[Spherical logarithmic mapping] Given the spherical feature $x\in\mathbb{S}_r^{d+1}$ with $\Vert x\Vert = r$, we define the spherical logarithmic transform \cite{DBLP:journals/pami/WilsonHPD14} $\log_N:\mathbb{S}^{d+1}_r\rightarrow T_N\mathbb{S}^{d+1}_r$ by
        \begin{equation}
            \log_N (x) = \frac{\psi}{\sin\psi}(x-N\cos\psi),
        \end{equation}
        where $N=(0,\cdots,0,r)\in\mathbb{S}_r^{d+1}$ is the north pole, $\psi = {\rm argcos}(N^Tx/r^2)$.
        Further, the spherical logarithmic mapping $\mathcal{LOG}:\mathbb{R}^{h'\times w'\times (d+1)}\rightarrow\mathbb{R}^{h'\times w'\times d}$ for a spherical feature map $\Phi$ is defined by
        \begin{equation}
            \mathcal{LOG}(\Phi)[i,j,:] = \mathcal{H}\left(\log_N(\Phi[i,j,:])\right),
        \end{equation}
        where $\mathcal{H}:T_N\mathbb{S}^{d+1}_r\rightarrow\mathbb{R}^d$ is defined by $\mathcal{H}((v,r))=v$.
    \end{definition}
    \begin{definition}[Spherical space distance]\label{def} Given two spherical feature maps $\Phi_1, \Phi_2 \in \mathbb{R}^{h'\times w'\times (d+1)}$ with $\Vert \Phi_1[i,j,:]\Vert=\Vert\Phi_2[i,j,:]\Vert=r$ for any $i,j$, the spherical space distance of $\Phi_1$ and $\Phi_2$ is defined as
        \begin{equation}\label{eq:sphericaldistance}
            \mathcal{S}\left\{\Phi_1, \Phi_2\right\} = \sum_{i=1}^{h'}\sum_{j=1}^{w'} 1-\frac{1}{r^2}\Phi_1[i,j,:]^T\Phi_2[i,j,:].
        \end{equation}
    \end{definition}

    \subsubsection{Encoder}\label{sec:encoder}
    We use the feature extraction of $D_{LR}$ as an example to explain the procedure, and $R$ can be carried out similarly to $D_{LR}$ by changing the subscripts from $D$ to $R$. First, a $3\times3$ convolution is used to obtain shallow features embedding $\Phi_{D}^{(0)}$.
    The main body of feature extraction consists of a cascade of $P$ Restormer blocks, and we denote the $p$-th Restormer block in $\mathcal{E_D}$ as $\mathcal{R}_D^{(p)}$, where $p =1,\cdots,P$.
    The input of each $\mathcal{R}_D^{(p)}$ is represented by $\Phi_{D}^{(p-1)}$.

    At the $p$-th step of feature extraction, $\Phi_{D}^{(p-1)}$ passes through $\mathcal{R}_D^{(p)}$ to obtain the preliminary extracted feature $\tilde \Phi_D^{(p)}$. According to the analysis for motivation in \cref{sec:1}, after the multi-head self-attention mechanism,  $\tilde \Phi_D^{(p)}$ should contain shared features in some channels and private features in other channels. Thus, we assume that features in the first $\frac{dim}{2}$ channels are shared and represent cross-modality information, while features in the latter $\frac{dim}{2}$ channels are private and represent the characteristics of their own modality.
    To achieve feature separation and alignment, $\tilde \Phi_D^{(p)}$ is mapped to the spherical space using the \textit{spherical exponential mapping} $\mathcal{EXP}(\cdot)$, and then recovered in the feature domain by the \textit{spherical logarithmic mapping} $\mathcal{LOG}(\cdot)$ after calculating the feature decomposition loss. Finally, the recovered features are re-concatenated along the channel dimension to obtain $\Phi_D^{(p)}$, which will be input to the next $\mathcal{R}_D^{(p+1)}$. The feature decomposition loss will be illustrated in subsequent sections. The total feature extraction process of $p$-th step is:
    \begin{subequations}\label{eq:update}
        \small
        \begin{align}
            \tilde \Phi_D^{(p)}        & =  \mathcal{R}_D^{(p)}\left(\Phi_{D}^{(p-1)}\right)                                                                                    \label{eq:updatesub1} \\
            \tilde\Phi_{D,(p)}^{align} & = \tilde\Phi_D^{(p)}\left[ 0:\tfrac{dim}{2} \right ],\tilde\Phi_{D,(p)}^{sepn} = \tilde\Phi_D^{(p)}\left [\tfrac {dim}{2}:dim\right ] \label{eq:updatesub2} \\
            \Phi_{D,(p)}^{align}  & = \mathcal{LOG} \left ( \mathcal{EXP} \left ( \tilde\Phi_{D,(p)}^{align} \right )   \right )                                           \label{eq:updatesub3} \\
            \Phi_{D,(p)}^{sepn}   & = \mathcal{LOG}\left ( \mathcal{EXP} \left ( \tilde\Phi_{D,(p)}^{sepn} \right )   \right )                                             \label{eq:updatesub4} \\
            \Phi_{D}^{(p)}             & = {Cat}\left (  \Phi_{D,(p)}^{align},  \Phi_{D,(p)}^{sepn} \right )\label{eq:updatesub5}
        \end{align}
    \end{subequations}
    where $\{\tilde\Phi_{D,(p)}^{align},\tilde\Phi_{D,(p)}^{sepn}\}$ and $\{ \Phi_{D,(p)}^{align}, \Phi_{D,(p)}^{sepn}\}$ are the aligned and separated features after/before calculating the feature decomposition loss, and ${Cat}(\cdot,\cdot)$ is the channel concatenation operator.

    Eventually, the entire encoder feature extraction can be regarded as:
    \begin{equation}\label{eq:encoder}
        \Phi_{D} = \mathcal{E}_{D}\left( D_{LR}\right),\ \Phi_{R} = \mathcal{E}_{R}\left( R\right),
    \end{equation}
    where $\Phi_{D}$ and $\Phi_{R}$ are abbreviated forms of $\Phi_{D}^{(P)}$ and $\Phi_{D}^{(P)}$.
    Then $\Phi_{D}$ and $\Phi_{R}$ will be used as the input to the decoder.

    \subsubsection{Decoder}
    According to \cref{sec:encoder}, we obtained features $\Phi_{D}$ and $\Phi_{R}$, each containing shared information across $\frac{dim}{2}$ channels and private information across the other $\frac{dim}{2}$ channels. The features that we consider helpful for the HR depth reconstruction task are the full depth features $\Phi_{D}$ and the shared RGB features $\Phi_{R}^{sepn}$. Therefore, we concatenate $\Phi_{D}$ and $\Phi_{R}^{sepn}$ in channel dimension and input them to $\mathcal{D_D}$ to obtain the reconstructed depth image $\hat{D}_{HR}$, and input $\Phi_{R}$ to $\mathcal{D_R}$ to obtain the reconstructed RGB image $\hat{R}$, which is:
    \begin{equation}\label{eq:update2}
        \resizebox{.9\hsize}{!}{$
            \hat{D}_{HR}\!=\!\mathcal{D_D}\left(Cat\!\left(\Phi_{D},\Phi_{R}\left[ 0\!:\!\tfrac{dim}{2}\right]\right)\right)\!,\ \hat{R}\!=\!\mathcal{D_R}\!\left(\Phi_{R}\right). $}
    \end{equation}

    \newcommand{\hhspace}{-0.1em}
\newcommand{\widthh}{0.325}
\begin{figure*}[t]
    \centering
    \begin{subfigure}{\widthh\linewidth}
        \centering
        \includegraphics[width=\linewidth]{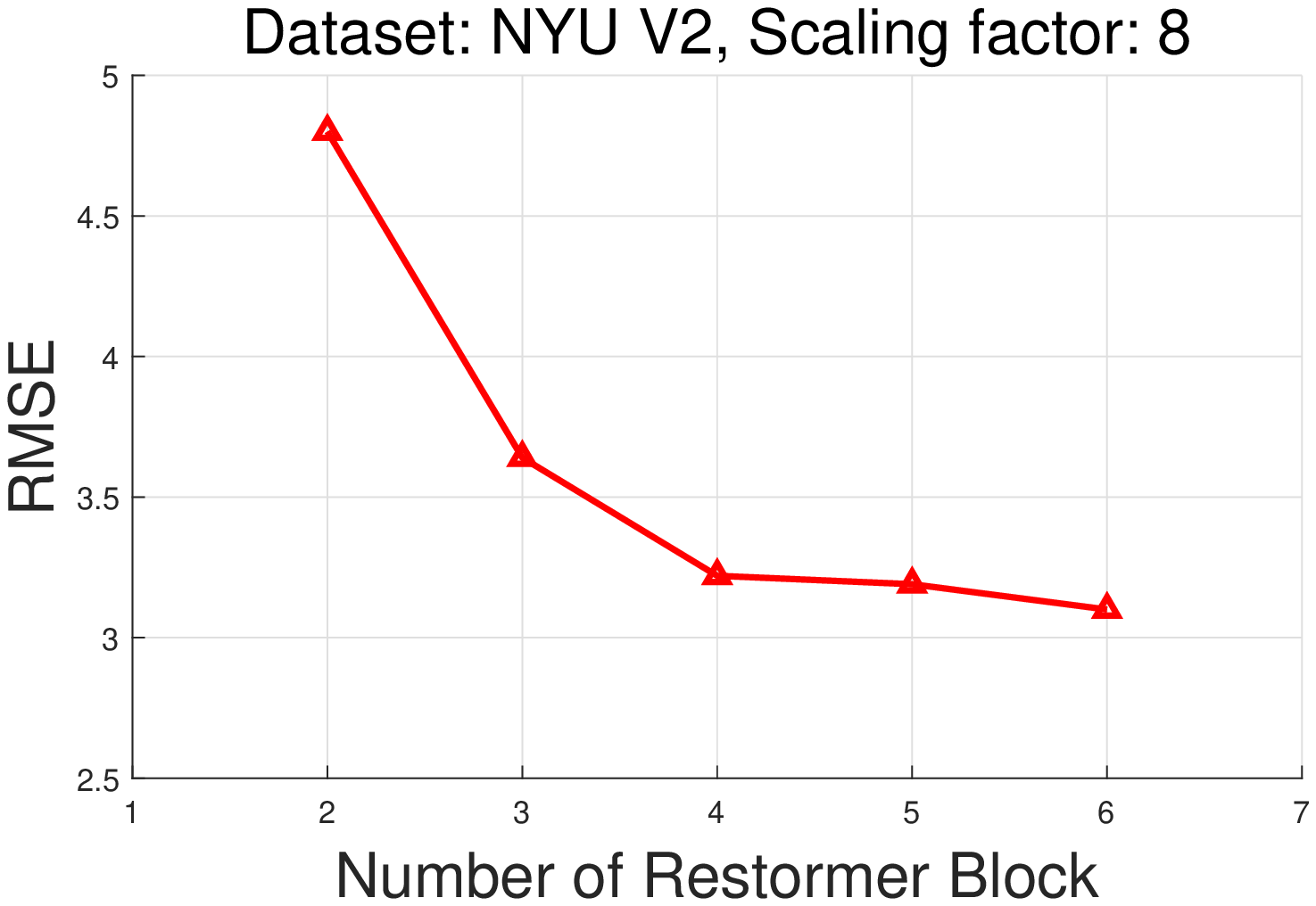}
    \end{subfigure}\hspace{\hhspace}
    \begin{subfigure}{\widthh\linewidth}
        \centering
        \includegraphics[width=\linewidth]{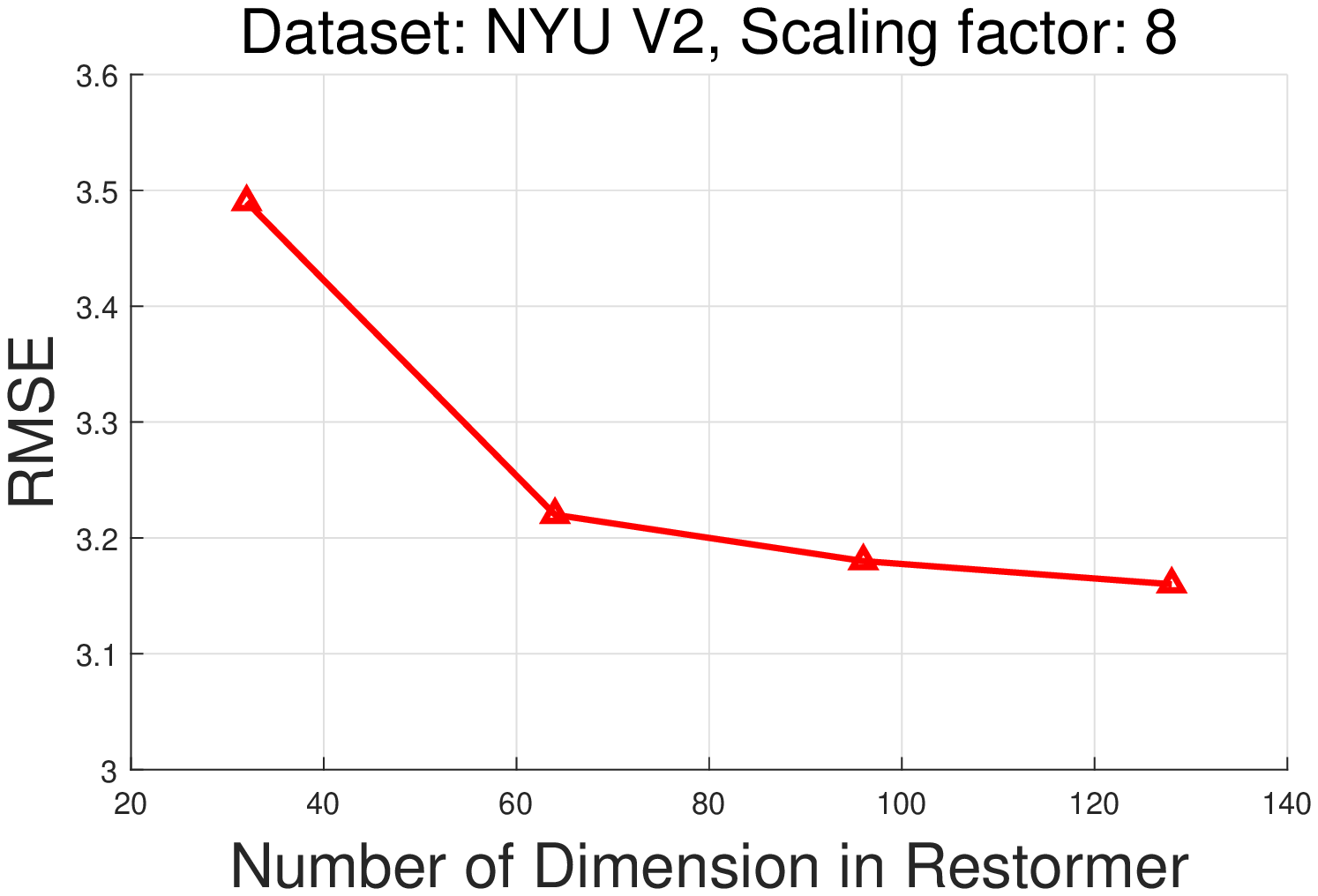}
    \end{subfigure}\hspace{\hhspace}
    \begin{subfigure}{\widthh\linewidth}
        \centering
        \includegraphics[width=0.97\linewidth]{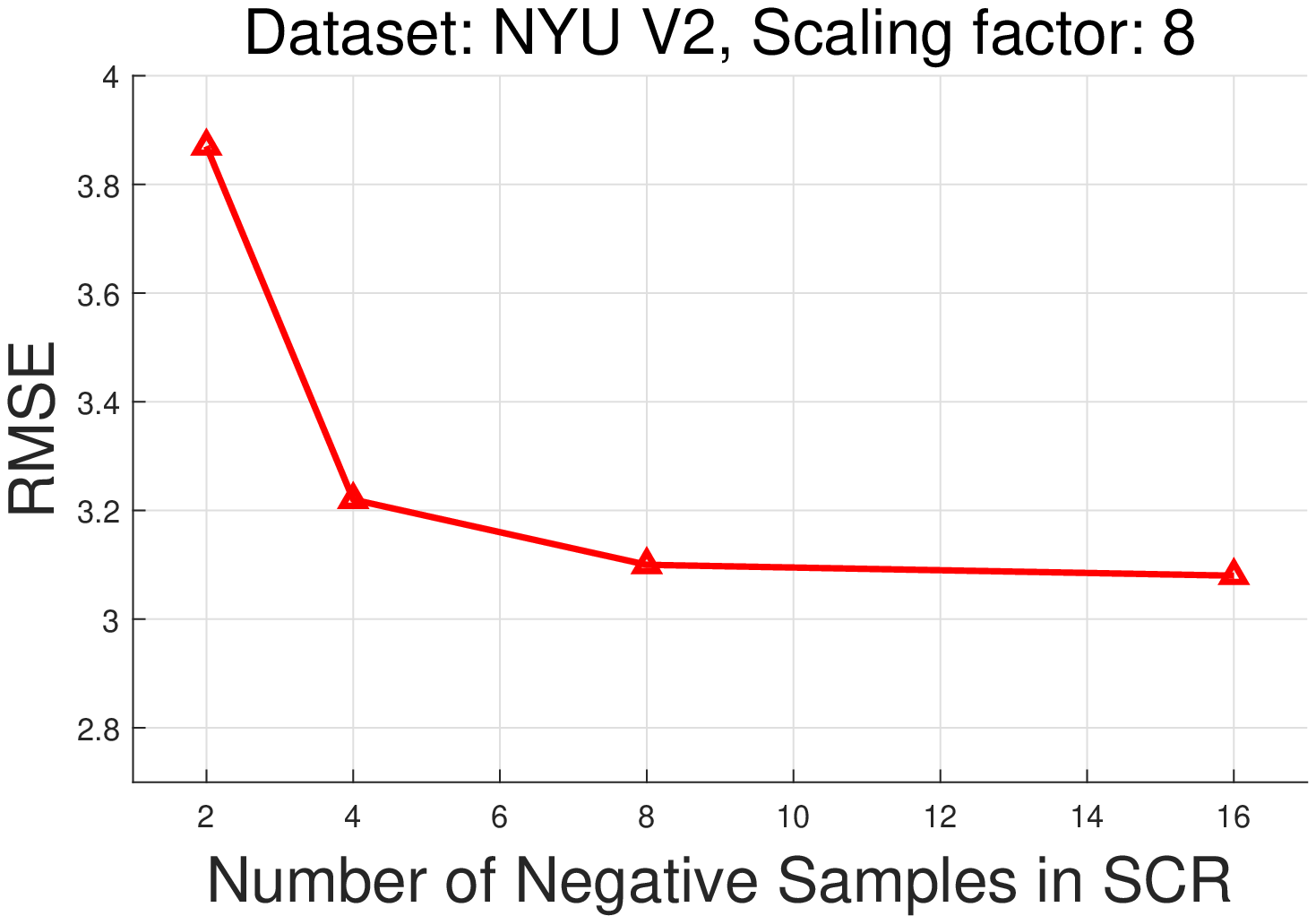}
    \end{subfigure}\hspace{\hhspace}
    \vspace{-0.2em}
    \caption{Exhibition for different settings of SSDNet. Results are evaluated on validation set of NYU V2 for GDSR $\times$ 8.}\label{fig:parameter}
    \label{fig:settings}
    \end{figure*}
    \begin{figure*}[t]
        \centering
        \includegraphics[width=0.8\linewidth]{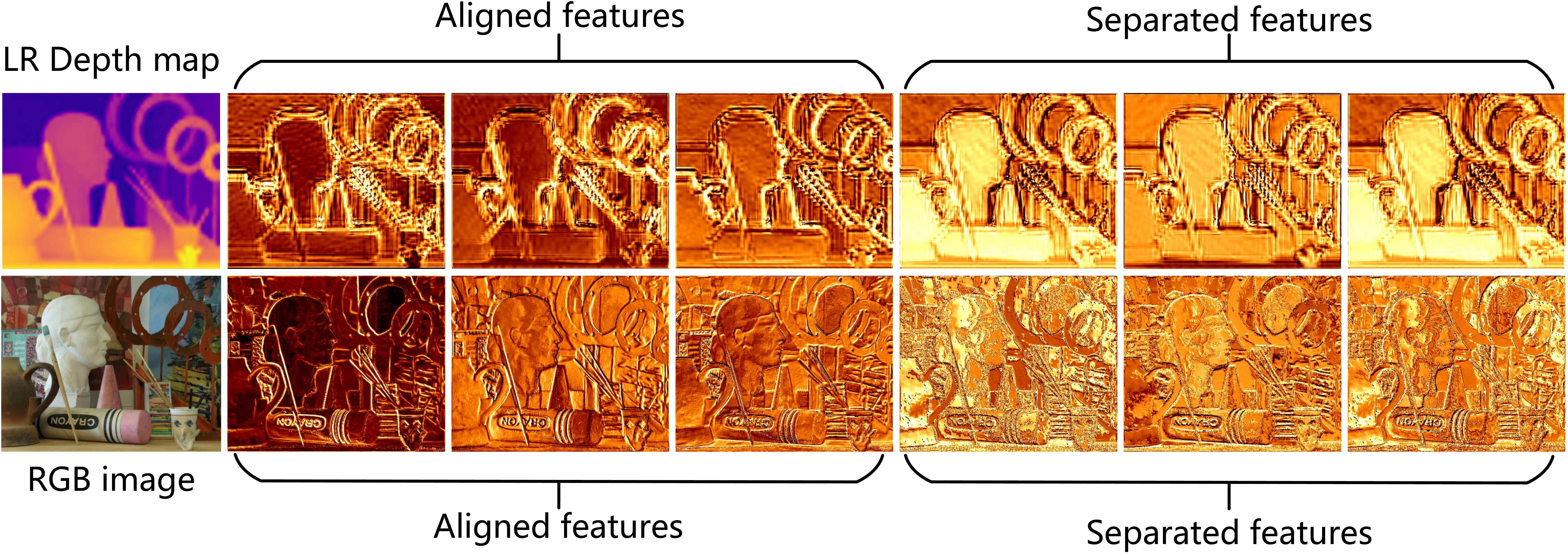}
        \caption{Visulization of the aligned and separated features.}
        \label{fig:rebuttal}
    \end{figure*}
    \begin{figure*}[t]
        \centering
        \includegraphics[width=0.975\linewidth]{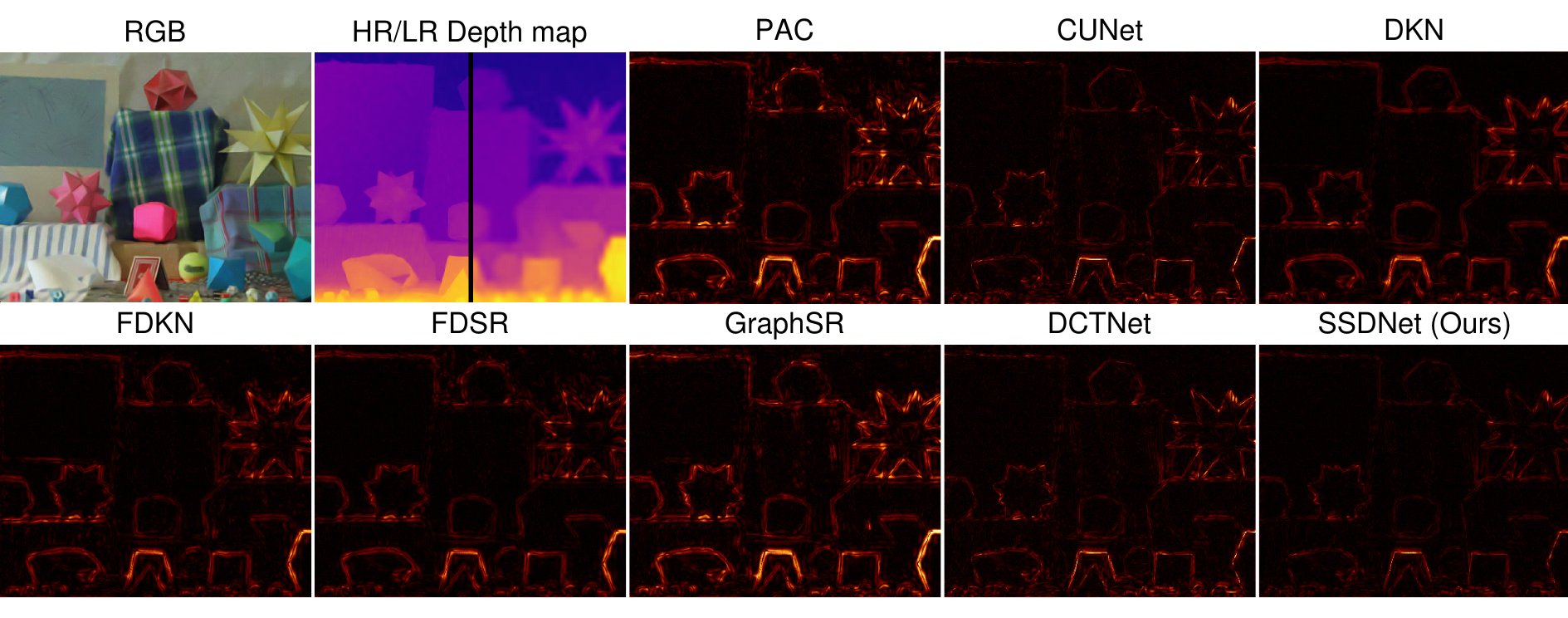}
        \caption{Visual error maps for test image in the Middlebury dataset for 8$\times$ super-resolution.}
        \label{fig:Visual1}
    \end{figure*}
    \begin{figure*}[t]
        \centering
        \includegraphics[width=0.975\linewidth]{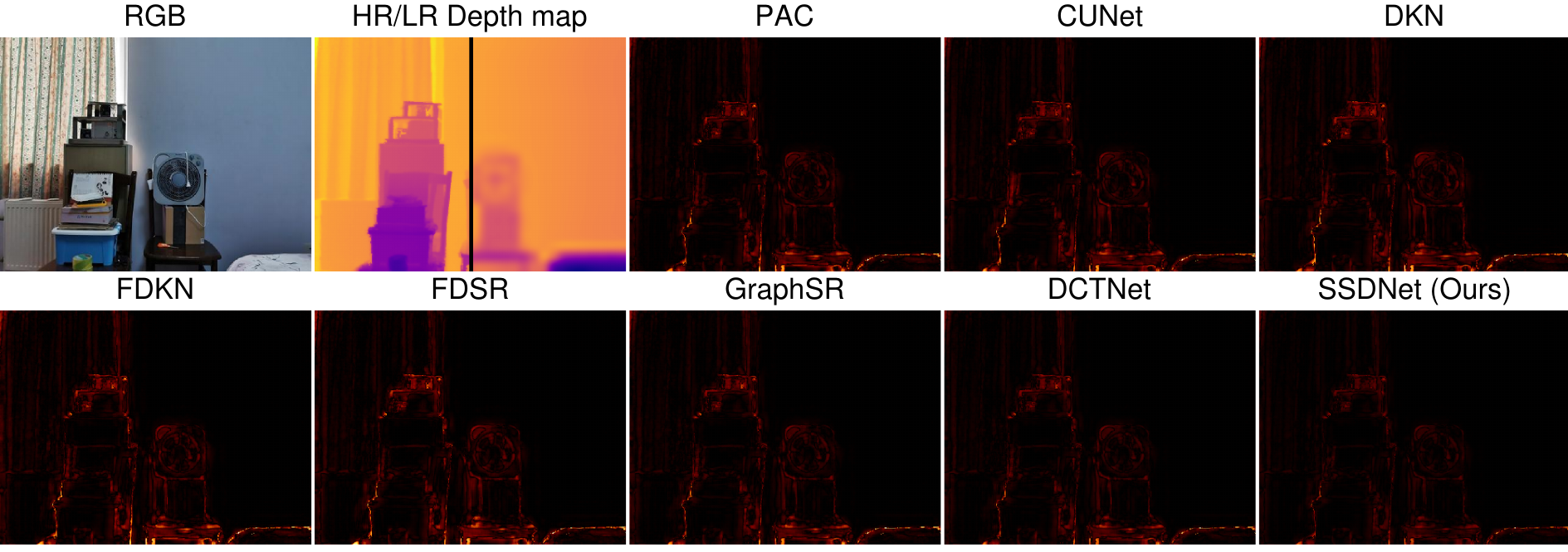}
        \caption{Visual error maps for test image in the RGBDD dataset for 16$\times$ super-resolution.}
        \label{fig:Visual2}
    \end{figure*}

    \subsubsection{Spherical Contrast Refinement module}
    After decoding, we have obtained a {preliminary} estimation of the HR depth map $\hat{D}_{H\!R}$. However, it potentially has some minor issues, such as blurry edges, noisy surfaces and over-transferred RGB texture. Hence, in the \textit{Spherical Contrast Refinement} (SCR) module, we target the imperfect issues in $\hat{D}_{H\!R}$ by contrastive learning, and the diagram is shown in \cref{fig:Workflow}\red{c}.

    \bfsection{Defect patches classifier}
    Firstly, we artificially synthesize the ``imperfect image dataset'' using the training set. For example, for an $m\times m$ image patch from the ground truth in the training set, we can add random noise to make the patch  \textit{noisy}, apply Gaussian blur to it  \textit{blurry}, and add the same-location RGB image to make it \textit{texture over-transferred}. Patches that are not processed can be regarded as \textit{perfect}. In this way, we obtain a dataset with labels ``noisy'', ``blurry'', ``texture over-transferred'' and ``perfect'' which can be used to train a \textit{defect patches classifier} (DPC) based on ResNet34~\cite{DBLP:conf/cvpr/HeZRS16}.

    \bfsection{Positive and negative samples}
    After obtaining the decoder output $\hat{D}_{H\!R}$ in \cref{eq:update2}, we randomly crop it into an $m\times m$ patch $\hat{D}_{H\!R}^{pat}$, and input it into the well-trained DPC to obtain the imperfect type of $\hat{D}_{H\!R}^{pat}$.
    After getting the imperfect label, similar to the operation for making the ``imperfect image dataset'', we can transform the corresponding ground truth ${D}_{H\!R}$ into an imperfect depth map $\tilde{D}_{H\!R}$. Then, we randomly crop $N$ $m\times m$ patches from $\tilde{D}_{H\!R}$ that are different from the position of $\hat{D}_{H\!R}^{pat}$, and the corresponding imperfect patches $\tilde{D}_{H\!R}^{pat}$, named \textit{negative samples}, are generated. Meanwhile, we refer to $\hat{D}_{H\!R}^{pat}$ as the \textit{anchor} and the same-location ground truth ${D}_{H\!R}^{pat}$ as the \textit{positive sample}. The $k$-th positive, anchor and negative samples are represented by ${\mu_k}^{+}$, ${\mu_k}$, and ${\mu_k}^{-}$, respectively.

    \bfsection{Spherical contrast refinement}
    We input ${\mu_k}^{+}$, ${\mu_k}$, and ${\mu_k}^{-}$ into $\mathcal{E}_{D}$ and get the contrastive refinement loss $\mathcal{L}_{SCR}$:
     \begin{equation}\label{eq:SCRloss}
         \mathcal{L}_{SCR} = \sum_{k=1}^{K}\frac{\mathcal{S}\left \{ \mathcal{E}_{D}\left({\mu_k}\right),\mathcal{E}_{D}\left({\mu_k}^{+}\right) \right \} }
         {\sum_{n = 1}^{N}{\mathcal{S}\left \{ \mathcal{E}_{D}\left({\mu_k}\right),\mathcal{E}_{D}\left({\mu_k}^{n-}\right) \right \} }},
     \end{equation}
     where $\mathcal{S}\left \{\cdot, \cdot\right \} $ is the spherical space distance in \cref{def}, and ${\mu_k}^{n-}$ is the $n$th negative sample in ${\mu_k}^{-}$.
     Through gradient descent, we minimize the distance between ${\mu_k}^{+}$ and ${\mu_k}$ features and maximize that between ${\mu_k}$ and ${\mu_k}^{-}$ features. This process fine-tunes $\mathcal{E}_{D}$. However, since the increased training cost for the SCR module, we incorporate it into the regular network training process every few iterations to strike a balance between training efficiency and effectiveness.

    \subsection{Training loss}
    The training loss in this paper comprises several components: the depth map reconstruction loss $\mathcal{L}_{pixel}^D$, the RGB reconstruction loss $\mathcal{L}_{pixel}^R$, the feature decomposition loss $\mathcal{L}_{dec}$, and the spherical contrastive refinement loss $\mathcal{L}_{SCR} $. We describe each loss separately next.

    $\mathcal{L}_{pixel}^D$ ensures that the estimated depth map $\hat{D}_{HR}$ output by our SSDNet is close to the ground truth depth map ${D}_{HR}$. $\mathcal{L}_{pixel}^R$ ensures that the output $\hat R$ is close to the input $R$. Although theoretically unrelated to the depth map super-resolution task, this loss item is used to guarantee that the semantic information from the RGB image is involved in the shared RGB features, rather than simply generating random noise that is approximate to the depth features to meet the feature decomposition. Specifically,
    \begin{equation}\label{eq:depthloss}
        \small
        \mathcal{L}_{pixel}^D\!=\!\sum_{k=1}^K \!\| \hat{D}_{HR}^{(k)}\!-\!{D}_{HR}^{(k)} \|_2^2,\ \mathcal{L}_{pixel}^R\!=\!\sum_{k=1}^K\!\| \hat{R}^{(k)}\!-\!R^{(k)} \|_2^2.
    \end{equation}
    Regarding the feature decomposition loss $\mathcal{L}_{dec}$, we utilize the \textit{spherical space distance} to enhance the similarity between shared features while reducing the similarity between separated features.
    We define the specific structure of the feature decomposition loss $\mathcal{L}_{dec}$ as follows:
    \begin{equation}\label{eq:decloss}
        \mathcal{L}_{dec} = {\mathcal{L}_{align}} - ({1-\mathcal{L}_{sepn}})^2,
    \end{equation}
    where
    \begin{equation}\label{eq:decloss2}
        \small
        \mathcal{L}_{sepn}\!=\!\sum_{p=1}^{P}\!\mathcal{S}\!\left\{\!\Phi_{D,(p)}^{sepn},\Phi_{R,(p)}^{sepn}\!\right\}, \mathcal{L}_{align}\!=\!\sum_{p=1}^{P}\!\mathcal{S}\left\{\!\Phi_{D,(p)}^{align} ,\Phi_{R,(p)}^{align}\!\right\}.
    \end{equation}
    Unlike $\ell_2$ distance, the spherical distance captures relative differences without being influenced by scale.
    For 1D feature maps with three consecutive pixels: $f_1\!=\![0.4,0.5,0.6]$, $f_2\!=\![4,5,6]$, $f_3\!=\![0.6,0.5,0.4]$, $f_1$ and $f_2$ exhibit similar pixel-wise increase structures and potentially similar extracted features, while $f_3$ differs. Thus, $\{f_1,f_2\}$ should be closer in distance and aligned, while $\{f_1,f_3\}$ should be separated. However, $\ell_2\!\left\{f_1,f_2\right\}\!=\!7.90$, $\ell_2\!\left\{f_1,f_3\right\}\!=\!0.28$ while $\mathcal{S}\{f_1,f_2\}\!=\!7.7\!\times\!10^{-8}$, $\mathcal{S}\{f_1,f_3\}\!=\!0.02$. Therefore, the spherical distance is a better choice.

    When optimizing $\mathcal{L}_{dec}$, since the value range of spherical distance $\mathcal{S}\{\cdot,\cdot\}$ is $[0, 2]$, we have $\mathcal{L}_{align}\!\to\!0$, $\mathcal{L}_{sepn}\!\to\!1$. According to \cref{def}, the derivation gives: $\mathcal{C}\{\Phi_{D,(p)}^{align} ,\Phi_{R,(p)}^{align}\}\!\to\!1$ and $\mathcal{C}\{\Phi_{D,(p)}^{sepn},\Phi_{R,(p)}^{sepn}\}\!\to\!0$, where $\mathcal{C}$ is the cosine similarity (the last term of \cref{eq:sphericaldistance}). When $\mathcal{C}\!=\!1$, vectors have an angle of 0, indicating higher similarity. If $\mathcal{C}\!=\!0$, the vectors are orthogonal, signifying no correlation. By forcing $\mathcal{C}\!=\!1$ or $\mathcal{C}\!=\!0$ for aligned/private features mapped to spherical space, we ensure the enhancement/reduction of similarity, which achieves our goal of feature decomposition.

    By combining \cref{eq:depthloss,eq:decloss,eq:SCRloss}, we obtain the total loss used for training, which can be expressed as follows:
    \begin{equation}\label{eq:totalloss}
        \mathcal{L}_{total} = \mathcal{L}_{pixel}^D + \alpha_1\mathcal{L}_{pixel}^R + \alpha_2\mathcal{L}_{dec} + \alpha_3\mathcal{L}_{SCR}.
    \end{equation}
    Note that $\alpha_3=0$ if the current epoch does not require the spherical contrast refinement.

\begin{table*}[t]
    \centering
    \resizebox{0.85\linewidth}{!}{
        \begin{tabular}{lcccccccccccc}
            \toprule
            \multirow{2}{*}{Methods}                                                    & \multicolumn{3}{c}{Middlebury~\cite{DBLP:conf/cvpr/HirschmullerS07,DBLP:conf/cvpr/ScharsteinP07}} & \multicolumn{3}{c}{NYU V2~\cite{DBLP:conf/eccv/SilbermanHKF12}} &  \multicolumn{3}{c}{Lu~\cite{DBLP:conf/cvpr/LuRL14}}   & \multicolumn{3}{c}{RGBDD~\cite{DBLP:conf/cvpr/HeZLBCZLL021}} \\
            \cmidrule(lr){2-4}\cmidrule(lr){5-7}\cmidrule(lr){8-10}\cmidrule(lr){11-13} &    $\times$4     &    $\times$8     &                         $\times$16                          &    $\times$4     &    $\times$8     &        $\times$16         &    $\times$4     &    $\times$8     &    $\times$16    &    $\times$4     &    $\times$8     &       $\times$16       \\ \midrule
            PAC~\cite{DBLP:conf/cvpr/SuJSGLK19}                                         &       1.32       &       2.62       &                            4.58                             &       1.89       &       3.33       &           6.78            &       1.20       &       2.33       &       5.19       &       1.25       &       1.98       &          3.49          \\
            CUNet~\cite{DBLP:journals/pami/0002D21}                                     &       1.10       &       2.17       &                            4.33                             &       1.92       &       3.70       &           6.78            &       0.91       &       2.23       &       4.99       &       1.18       &       1.95       &          3.45          \\
            DKN~\cite{DBLP:journals/ijcv/KimPH21}                                       &       1.23       &       2.12       &                            4.24                             &       1.62       &       3.26       &           6.51            &       0.96       &       2.16       &       5.11       &       1.30       &       1.96       &          3.42          \\
            FDKN~\cite{DBLP:journals/ijcv/KimPH21}                                      & \underline{1.08} &       2.17       &                            4.50                             &       1.86       &       3.58       &           6.96            & \underline{0.82} &       2.10       &       5.05       &       1.18       &       1.91       &          3.41          \\
            FDSR~\cite{DBLP:conf/cvpr/HeZLBCZLL021}                                     &       1.13       &       2.08       &                            4.39                             &       1.61       &       3.18       &          {5.86}           &       1.29       &       2.19       &       5.00       &       1.16       &       1.82       &          3.06          \\
            GraphSR~\cite{DBLP:conf/cvpr/LutioBDRWS22}                                  &       1.11       &       2.12       &                            4.43                             &       1.79       &       3.17       &           6.02            &       0.92       &       2.05       &       5.15       &       1.30       &       1.83       &          3.12          \\
            DCTNet~\cite{DBLP:conf/cvpr/ZhaoZXLP22}                                     &       1.10       & \underline{2.05} &                      \underline{4.19}                       &  \textbf{1.59}   & \underline{3.16} &       \textbf{5.84}       &       0.88       & \underline{1.85} &  \textbf{4.39}   & \underline{1.08} & \underline{1.74} &    \underline{3.05}    \\
            Ours                                                                        &  \textbf{1.02}   &  \textbf{1.91}   &                        \textbf{4.02}                        & \underline{1.60} &  \textbf{3.14}   &     \underline{5.86}      &  \textbf{0.80}   &  \textbf{1.82}   & \underline{4.77} &  \textbf{1.04}   &  \textbf{1.72}   &     \textbf{2.92}      \\ \bottomrule
        \end{tabular}}
    \caption{Quantitative comparisons among the SOTA methods and our SSDNet in test datasets. The best and second best values are highlighted by \textbf{bold} and \underline{underline}, respectively.}
    \label{tab:quant}
\end{table*}

\begin{table}[t]
    \centering
    \resizebox{\linewidth}{!}{
        \begin{tabular}{lclclc}
            \toprule
            \multicolumn{6}{c}{\textbf{Dataset: RGBDD in real-world}}                                                                    \\
            Methods                                     &       RMSE       & Methods                                     &     RMSE      & Methods                                 &     RMSE      \\ \midrule
            SVLRM~\cite{DBLP:conf/cvpr/PanDRLT019}      &       8.05       & DJF~\cite{DBLP:conf/eccv/LiHA016}           &     7.90      & DJFR~\cite{DBLP:journals/pami/LiHAY19}  &     8.01      \\
            DKN~\cite{DBLP:journals/ijcv/KimPH21}       &      {7.38}      & FDKN~\cite{DBLP:journals/ijcv/KimPH21}      &     7.50      & FDSR~\cite{DBLP:conf/cvpr/HeZLBCZLL021} &     7.50      \\
            DCTNet~\cite{DBLP:conf/cvpr/ZhaoZXLP22}     & \underline{7.37} & SSDNet                                      & \textbf{7.32} &                                         &               \\\midrule
            FDSR$^*$~\cite{DBLP:conf/cvpr/HeZLBCZLL021} &       5.49       & DCTNet$^*$~\cite{DBLP:conf/cvpr/ZhaoZXLP22} &    {5.43}     & SSDNet$^*$                              & \textbf{5.38} \\ \bottomrule
    \end{tabular}}
    \caption{Quantitative comparisons among the SOTA methods and SSDNet on \textit{real-world branch} of RGBDD. The best and second best RMSEs are highlighted by \textbf{bold} and \underline{underline}. Model$^*$ means that the model has been fine-tuned on the real-world training data.}
    \label{tab:quant2}
\end{table}
    \begin{table}[t]
        \centering
        \resizebox{\linewidth}{!}{
            \begin{tabular}{ccccccccc}
                \toprule
                Methods & PAC   & CUN   & DKN   & FDKN  & FDSR  & GSR   & DCT   & Ours \\\midrule
                Time (s)  & 0.20  & 0.23  & 0.21  & 0.04  & 0.01  & 0.92  & 0.08  & 0.10 \\\bottomrule
        \end{tabular}}
        \caption{Results of time-consuming comparison for generating a 640$\times$ 480 HR depth map.}
        \label{tab:rebuttal2}%
    \end{table}%
    \begin{table*}[t]
        \centering
        \resizebox{0.8\linewidth}{!}{
            \begin{tabular}{lcccccccc}
                \toprule
                & Encoder\&Decoder & $\mathcal{L}_{dec}$ & Form of $\mathcal{L}_{dec}$ & $\mathcal{L}_{pixel}^R$ &   SCR Module  &   $\times$4   &   $\times$8   &  $\times$16 \\ \midrule
                Exp.~\uppercase\expandafter{\romannumeral 1} &      Shared      &       $\surd$       & $\mathcal{S}$ &         $\surd$         & $\surd$   &     1.23      &     2.10      &     3.49 \\
                Exp.~\uppercase\expandafter{\romannumeral 2} &     Private      &      $\times$       & $\mathcal{S}$ &         $\surd$         & $\surd$ &     1.17      &     1.94      &     3.27  \\
                Exp.~\uppercase\expandafter{\romannumeral 3} &     Private      &       $\surd$       &   $\ell_2$    &         $\surd$         & $\surd$   &     1.15      &     1.83      &     3.06 \\
                Exp.~\uppercase\expandafter{\romannumeral 4} &     Private      &       $\surd$       & $\mathcal{S}$ &        $\times$         & $\surd$  &     1.16      &     1.89      &     3.29 \\
                Exp.~\uppercase\expandafter{\romannumeral 5} &     Private      &       $\surd$       & $\mathcal{S}$ &         $\surd$         & $\times$ &     1.18      &     1.87      &     3.36 \\
                \midrule
                Ours         &     Private      &       $\surd$       & $\mathcal{S}$ &         $\surd$         & $\surd$ & \textbf{1.04} & \textbf{1.72} & \textbf{2.92} \\ \bottomrule
        \end{tabular}}
    \caption{Results of ablation experiments on the RGBDD test set. \textbf{Bold} indicates the best score in terms of RMSE.}
    \label{tab:Ablation}
    \end{table*}%

    \section{Experiment}\label{sec:4}
    This section will conduct a comprehensive set of experiments that aim to showcase the effectiveness of our model in addressing the GDSR task. Moreover, we will also provide evidence that substantiates the soundness of our SSDNet.
    \subsection{Setup} \label{sec:setup}
    \bfsection{Datasets}
    Our experiments follow the protocol established in \cite{DBLP:conf/cvpr/ZhaoZXLP22,DBLP:conf/cvpr/HeZLBCZLL021,DBLP:journals/ijcv/KimPH21}. To evaluate our model, we employ four popular benchmarks: Middlebury~\cite{DBLP:conf/cvpr/HirschmullerS07,DBLP:conf/cvpr/ScharsteinP07}, Lu~\cite{DBLP:conf/cvpr/LuRL14}, NYU v2~\cite{DBLP:conf/eccv/SilbermanHKF12}, and RGBDD~\cite{DBLP:conf/cvpr/HeZLBCZLL021}. Our training and validation sets consist of the first 1000 images of NYU v2 dataset~\cite{DBLP:conf/eccv/SilbermanHKF12}, divided into a 9:1 ratio. The last 449 pairs in NYU v2 are utilized as the test dataset. Middlebury~\cite{DBLP:conf/cvpr/HirschmullerS07,DBLP:conf/cvpr/ScharsteinP07} (30 pairs), Lu~\cite{DBLP:conf/cvpr/LuRL14} (6 pairs), and RGBDD~\cite{DBLP:conf/cvpr/HeZLBCZLL021} (405 pairs) are also used as test sets to evaluate the depth map SR ability across different scenes and objects.
    To furthermore demonstrate the generalization ability of our model to unknown scenarios, we incorporate the\textit{real-world branch} of the RGBDD dataset, which comprises 2215/405 pairs of RGB-D images for training/test sets, respectively. Lastly, we use the root-mean-square error~(RMSE) metric to evaluate the super-resolution effect of our model.

    \bfsection{Implementation details}
    In our experiments, we apply bicubic down-sampling to the HR depth maps to synthesize the corresponding LR depth maps.
    During the pre-processing stage, the training samples undergo resizing to 128$\times$128. We train the network with a mini-batch size of 56 for 100 epochs, employing the Adam~\cite{kingma2014adam} optimizer with an initial learning rate of $5\times10^{-3}$. For training and testing, we employ a PC featuring eight NVIDIA GeForce RTX 3090 GPUs.
    $\alpha_1$, $\alpha_2$ and $\alpha_3$ in \cref{eq:totalloss} set to $10^{-2}$, $10^{-3}$ and $10^{-2}$ in order to balance the magnitudes of each term in the loss. The SCR fine-tuning is performed once every 10 epochs of training. Further details are provided in the \textit{supplementary material}.

    \subsection{Empirical validation experiment}
    \bfsection{Impact of network hyper-parameters}
    In our proposed SSDNet, the number of Restormer Blocks $P$, and number of dimensions in Restormer $C$, and the number of negative samples in SCR $N$ are crucial factors in enhancing the super-resolution performance.
    We conduct experiments on the NYU validation set to study the impact of different combinations of ${P,C,N}$. Initially, we fix any two parameters at $C=64$, $P=4$, and $N=8$, respectively, and evaluate the prediction quality of the unfixed parameter at $P=2,3,4,5,6$, $C=32,64,96,128$, and $N=2,4,8,16$. The results are summarized in \cref{fig:settings}. We observe that the performance is limited when $P<4$, and redundant parameter increases do not yield corresponding improvements when $P>4$. Similarly, increasing $C$ to more than 64 does not produce significant benefits but instead increased the training cost and computational expense. For the SCR module, the increase in training burden does not match the improvement in effectiveness when $N$ exceeds 8. Therefore, to ensure a balance between performance and computational cost, we choose $P=4, C=64, N=8$ for subsequent experiments.

    \bfsection{Visualization of feature decomposition} Aligned and separated features are visualized in \cref{fig:rebuttal}. Aligned features extract shared properties (edges/contours) from the depth/RGB image pair.
    Separated features capture modality-specific details (textures in RGB images, smooth depth distance in depth maps). The visualization aligns with our motivation.

    \subsection{Comparison with SOTA methods}
    This section aims to evaluate the performance of our SSDNet on several popular benchmarks in \cref{sec:setup}, and compare our results with the state-of-the-art methods, including PAC~\cite{DBLP:conf/cvpr/SuJSGLK19},
    CUNet~\cite{DBLP:journals/pami/0002D21},
    DKN~\cite{DBLP:journals/ijcv/KimPH21},
    FDKN~\cite{DBLP:journals/ijcv/KimPH21},
    FDSR~\cite{DBLP:conf/cvpr/HeZLBCZLL021},
    GraphSR~\cite{DBLP:conf/cvpr/LutioBDRWS22}, and
    DCTNet~\cite{DBLP:conf/cvpr/ZhaoZXLP22}.

    \bfsection{Qualitative Comparison}
    We present comparisons of the error maps with ground truth depth maps among SOTA methods in  \cref{fig:Visual1,fig:Visual2}. Intuitively, the depth predictions of SSDNet exhibit lower prediction errors and are closer to the ground truth images, especially in terms of perception and restoration for the edges and contours. Further visual comparisons are shown in the supplementary material.

    \bfsection{Quantitative results for traditional testsets}
    The quantitative results with scaling factors of $4$, $8$, and $16$ on four test sets are shown in \cref{tab:quant}. Compared with existing methods that achieve good results on certain datasets or super-resolution factors, our SSDNet produces satisfactory predictions on multiple datasets and various super-resolution scales.  This indicates that our model has a better super-resolution performance than previous methods.

    \bfsection{Quantitative results for real-world branch}
    In addition, following \cite{DBLP:conf/cvpr/HeZLBCZLL021,DBLP:conf/cvpr/ZhaoZXLP22}, we use the pre-trained $4\times$ model to directly test on the \textit{real-world branch} of the RGBDD dataset to explore the generalization ability in real-world scenarios. We conduct all testing without any additional fine-tuning, and the quantitative results are presented in \cref{tab:quant2}. Furthermore, we also perform targeted training and testing on this dataset. Regardless of whether fine-tuning or not, SSDNet outperforms previous methods in RMSE, highlighting its powerful generalization ability in handling unknown scenes.

    \bfsection{Parameter \& running time comparison}
    We demonstrate the relationship between the number of model learnable parameters \vs RMSE on the RGBDD dataset for $\times$4, $\times$8, and $\times$16 and Middlebury for $\times$4 in \cref{fig:introduction}.
    Our model exhibits a clear advantage over existing methods with relatively fewer parameters. The time-consuming comparison is presented in \cref{tab:rebuttal2}. They both demonstrate the efficiency of our method for the GDSR task, showing the potential for developing lightweight networks and practical applications in the future.
    \subsection{Ablation Studies}
    In this section, we validate the design rationality of our SSDNet through ablation experiments based on the RGBDD testset, and present the results in \cref{tab:Ablation}.

    \bfsection{SSDNet architecture}
    In Exp.~\uppercase\expandafter{\romannumeral1}, we share the encoders and decoders for RGB and depth images, \ie, a unified encoder and decoder are used instead of $\left\{\mathcal{E}_{D},\mathcal{E}_{R}\right\}$ and $\left\{\mathcal{D}_{D},\mathcal{D}_{R}\right\}$. The number of Restormer blocks is increased to ensure that the learnable parameter number is comparable.

    \bfsection{Spherical feature decomposition}
    In Exp.~\uppercase\expandafter{\romannumeral2}, we eliminate the entire feature decomposition module, \ie, $\mathcal{L}_{dec}$ in \cref{eq:decloss} will not be employed.
    In Exp.~\uppercase\expandafter{\romannumeral3}, we change the distance measure for $\mathcal{L}_{sepn}$ and $\mathcal{L}_{align}$ in \cref{eq:decloss2} from spherical space distance to Euclidean distance, \ie, $\ell_2$-loss.

    \bfsection{Reconstruction RGB}
    In Exp.~\uppercase\expandafter{\romannumeral4}, We eliminate $\mathcal{L}_{pixel}^R$ in \cref{eq:depthloss} and explore the change for feature extraction capability without the restriction of reconstructing RGB.

    \bfsection{Spherical contrast refinement}
    In Exp.~\uppercase\expandafter{\romannumeral5}, we remove the SCR module and use $\hat{D}_{HR}$ in \cref{eq:update2} as the final output of our model.

    \bfsection{Analysis}
    \cref{tab:Ablation} shows that changing settings leads to performance degradation, validating our model design.\\
    {Exp.~\uppercase\expandafter{\romannumeral 1}}: Unified encoder/decoder ignores modality-specific features, hindering efficient cross-modal feature extraction and causing the largest degradation.\\
    {Exp.~\uppercase\expandafter{\romannumeral 2}}: The absence of $\mathcal{L}_{dec}$ prevents decomposing the shared/private features.\\
    {Exp.~\uppercase\expandafter{\romannumeral 3}}: $\ell_2$ distance partially decomposes features but is less effective than $\mathcal{S}$.\\
    {Exp.~\uppercase\expandafter{\romannumeral 4}}: $\mathcal{L}_{pixel}^R$ preserves semantic information in RGB and avoids excessive adaptation of feature separation/alignment.\\
    {Exp.~\uppercase\expandafter{\romannumeral 5}}: Removal of SCR leads to mentioned detailed issues and performance degradation.

    \section{Conclusion}\label{sec:5}
    We propose a Spherical Space feature Decomposition network (SSDNet) for guided depth map super-resolution,
    where a Restormer-based encoder is used to extract the global features of the inputs, and the intermediate features will be mapped to the spherical space to complete the separation/alignment of shared/private features, respectively. Finally, a Restormer-based decoder is to reconstruct the HR depth map. The spherical contrast refinement module is then employed to address the possible detail issues, \eg, blurry edges, noisy surfaces and over-transferred RGB texture. The satisfactory output results and the lightweight size demonstrate the superiority of our approach.

    \section*{Acknowledgement}\label{sec:6}
    This work has been supported by the National Key Research and Development Program of China under grant 2018AAA0102201, the National Natural Science Foundation of China under Grant 61976174 and 12201497, Shaanxi Fundamental Science Research Project for Mathematics and Physics under Grant 22JSQ033, the Fundamental Research Funds for the Central Universities under Grant D5000220060, and partly supported by the Alexander von Humboldt Foundation.

    {\small
        \bibliographystyle{ieee_fullname}
        \bibliography{egbib}
    }

\end{document}